\documentclass[letterpaper, 10 pt, conference]{ieeeconf}  % Comment this line out if you need a4paper
\IEEEoverridecommandlockouts                              % This command is only needed if 
\usepackage{color}
\usepackage[utf8]{inputenc}
\usepackage{amsmath,amssymb}
\usepackage{mathtools}
\usepackage{amsthm}
\usepackage{graphicx}
\usepackage{float}
\usepackage{wrapfig}
\usepackage{algorithmicx,algpseudocode}
\usepackage{algorithm}
\usepackage{amsfonts,dsfont}
\usepackage{mathrsfs}  
\usepackage{booktabs}
\usepackage{hyperref}
\usepackage{cite}
\usepackage{balance} % To even out the references spread on two columns of a separate page

\renewcommand{\phi}{\varphi}
\renewcommand{\epsilon}{\varepsilon}

\usepackage{hybridsystem}

\DeclareMathOperator*{\argmin}{argmin}

\newcommand{\fmatrix}[1]
{\begin{bmatrix} 
f{#1}^{r}_v
(x #1)\\ 
f{#1}^{s}_v
(x #1) \end{bmatrix}}
\newcommand{\gmatrix}[1]{\begin{bmatrix} 
g{#1}^{r}_{1, v}
(x {#1}) & g{#1}^{r}_{2, v}
(x {#1}) \\
    0   
    & g{#1}^{s}_v
    (x {#1})   \end{bmatrix}}

\newcounter{thm}
\newcounter{lem}
\newcounter{cor}
\newcounter{prop}
\newcounter{defn}
\newcounter{rem}
\title{\LARGE \bf 
Emulating Human Kinematic Behavior on Lower-Limb Prostheses via Multi-Contact Models and Force-Based Nonlinear Control}
\author{Rachel Gehlhar and Aaron D. Ames
\thanks{*This material is based upon work supported by NSF Awards 1923239 and 1924526, and Wandercraft under Award No.
WANDERCRAFT.21. This research was approved by California Institute of Technology Institutional Review Board with protocol no. 21-0693 for human subject testing.}
\thanks{R. Gehlhar and A. Ames are with the Department of Mechanical and Civil Engineering, California Institute of Technology, Pasadena, CA 91125 USA. Emails:
{\tt\small $\{$rgehlhar, ames$\}$@caltech.edu}}
}

\begin{document}

\maketitle

%%%%%%%%%%%%%%%%%%%%%%%%%%%%%%%%%%%%%%%%%%%%%%%%%%%%%%%%%%%%%%%%%%%%%%%%%%%%%%%%
\begin{abstract}
Ankle push-off largely contributes to limb energy generation in human walking, leading to smoother and more efficient locomotion. Providing this net positive work to an amputee requires an active prosthesis, but has the potential to enable more natural assisted locomotion. 
To this end, this paper uses multi-contact models of locomotion together with force-based nonlinear optimization-based controllers to achieve human-like kinematic behavior, including ankle push-off, on a powered transfemoral prosthesis for 2 subjects. 
In particular, we leverage model-based control approaches for dynamic bipedal robotic walking to develop a systematic method to realize human-like walking on a powered prosthesis that does not require subject-specific tuning.
We begin by synthesizing an optimization problem that yields gaits that resemble human joint trajectories at a kinematic level, and realize these gaits on a prosthesis through a control Lyapunov function based nonlinear controller that responds to real-time ground reaction forces and interaction forces with the human. 
The proposed controller is implemented on a prosthesis for two subjects without tuning between subjects, emulating subject-specific human kinematic trends on the prosthesis joints. 
These experimental results demonstrate that our force-based nonlinear control approach achieves better tracking of human kinematic trajectories than traditional methods. 
\end{abstract}

%%%%%%%%%%%%%%%%%%%%%%%%%%%%%%%%%%%%%%%%%%%%%%%%%%%%%%%%%%%%%%%%%%%%%%%%%%%%%%%%

\section{Introduction}
In human walking, ankles contribute the most positive work of trailing leg joints in forward rocking \cite{soo2010mechanics} and contribute up to 60\% of the total energy generated by a limb during a gait cycle \cite{fuscaldi2008effects}.
Ankle push-off specifically contributes to the forward acceleration of the body \cite{Inman1966HumanL} and also greatly smooths the transition from double support to swing phase in human gait \cite{HumanWalking}. 
Researchers showed for a simple powered walking model, that toe push-off can supply energy to the system at a quarter of the cost of applying a hip torque because this toe push-off reduces the collision energy loss at heel strike \cite{kuo2001energetics}. For amputees specifically, increase in prosthetic ankle push-off reduces the loading impulse of the intact limb and the risk of knee osteoarthritis for amputees \cite{morgenroth2011effect}. While the importance of an ankle that can generate net positive work is clear, especially for amputees, current commercially available devices for amputees remain limited to passive devices. Additionally, for powered prosthesis controlling the ankle actuators to produce natural and dynamic walking has proven challenging. 

\begin{figure} [t]  
\centering
\includegraphics[width=1\columnwidth]{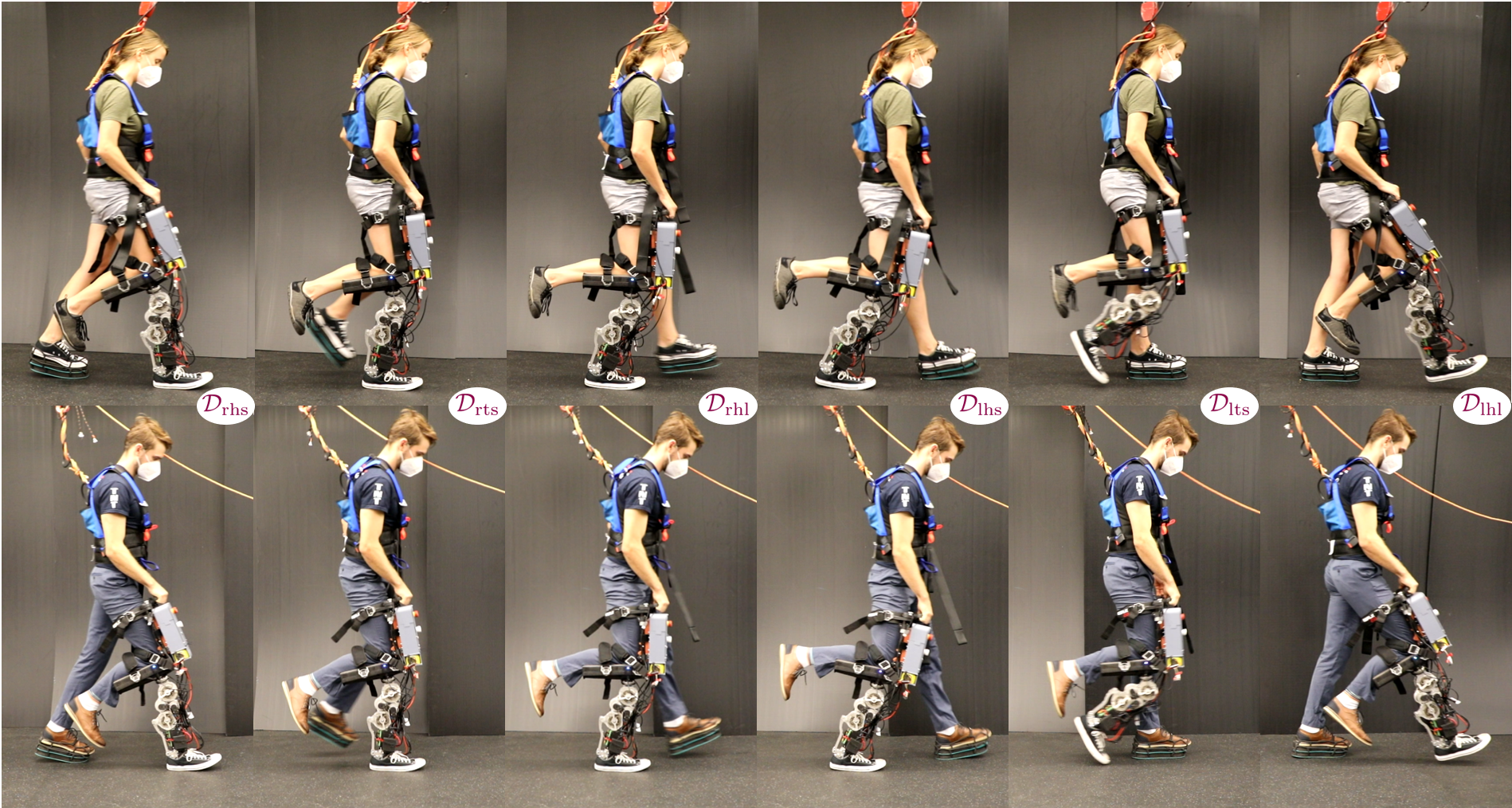}
\vspace{-0.6cm}
\caption{(top) Subject 1 and (bottom) Subject 2 walking with powered prosthesis controlled by multi-domain model-based prosthesis controller with real-time force sensing following provably stable human-like walking trajectories. Labels indicate the respective domain the human-prosthesis is in in the multi-domain hybrid-system graph shown in Fig. \ref{fig:hybrid-system}}
\label{fig:subjects}
\vspace{-0.6cm}
\end{figure}

Lower-limb passive prostheses cannot contribute net positive work to a human the way muscles do \cite{devita2007muscles}. Amputees walking with these devices expend more energy in walking, exhibit a decrease in comfortable walking speed, and have shorter step length \cite{WATERS1999207, Gailey_2008}. Powered prostheses can generate net positive work, but require a control strategy. One of the most commonly used control methods for these devices is impedance control \cite{DesignControlTransProsth, GoldfarbStair, bhakta2019impedance}, which divides the gait cycle into multiple discrete phases and tunes impedance parameters for each phase. This process takes hours to tune \cite{AnnSimonConfig5} to achieve human-like walking behavior \cite{goldfarb2014robotic} and needs to be repeated for every subject and behavior. Given there are over 600,000 lower-limb amputees in the US alone \cite{graham2008estimating}, it is not feasible to conduct this process for every potential user. 

With the goal of developing a more systematic method to construct controllers for powered prostheses, we look to model-based control methods developed for bipedal robots that generate stable walking trajectories offline \cite{sreenath2011compliant, ames2014rapidly, Reher2020algorithmic}.
The hybrid zero dynamics (HZD) framework \cite{westervelt2018feedback} these methods use accounts for the impact dynamics present in walking at foot strike and the zero dynamics that occur from underactuation. This framework also allows modeling of multiple domains of continuous dynamics and discrete connections between them as encoded by a directed cycle. This multi-domain hybrid system can model multi-contact behavior, and the changing contact behavior of heel-toe roll used for ankle push-off can be used to generate trajectories that satisfy formal guarantees of stability.
By tracking these trajectories online through a model-independent method, the end result has been multi-contact walking on bipedal robots \cite{Reher2020algorithmic, zhao2017multi} and a powered prosthesis \cite{zhao2016multi}. 

In order to retain stability guarantees online, control Lyapunov functions (CLFs) were employed in a quadratic program on bipedal robots \cite{galloway2015torque, reher2020inverse}. CLFs require knowledge of the full-order dynamics which is not available on a prosthesis since the human dynamics are unknown. Previous work developed a separable subsystem framework \cite{gehlhar2019control, StableRobustHZD} and proved that signals from an IMU and a load cell at the socket interface could complete the prosthesis model and full-order system stability could be guaranteed through a class of prosthesis subsystem controllers \cite{gehlhar2021separable}.

This paper implements a controller of the class in \cite{gehlhar2021separable} to realize multi-contact behavior on a prosthesis following a human-like trajectory generated through HZD methods. This paper differs from \cite{zhao2016multi} in both method and results, using a formally grounded force-based controller to achieve human-like walking, verified through the kinematics. 
The work of \cite{gehlhar2022powered} realized the first model-based lower-limb prosthesis controller with real-time force sensing on the knee joint of a powered prosthesis in 2 phases, stance and swing, with 1 prescribed walking gait.
This paper extends the work of \cite{gehlhar2022powered} in 3 major ways:
\begin{enumerate}
    \item we simultaneously apply the first model-based lower-limb prosthesis controller with real-time force sensing to the ankle in addition to the knee of a transfemoral prosthesis,
    \item we expand this model-based controller to a 6-domain hybrid system framework to emulate human multi-contact behavior (heel-toe roll),
    \item we demonstrate this new controller on 2 subjects with 2 different subject-specific prescribed walking gaits, and compare it to two other control methods.
\end{enumerate}

This paper overviews the multi-domain hybrid system used to model multi-contact behavior and presents the system dynamics used to model the human and prosthesis in Section \ref{sec:model}. Section \ref{sec:trajectories} describes the separable output functions used where we can use a subset of these outputs to construct a controller for the prosthesis. This section also covers the trajectory generation method that aims to match human kinematic data while also satisfying stability guarantees for the human-prosthesis system. To implement these trajectories, we construct our control method in Section \ref{sec:control}. In Section \ref{sec:experiment} we describe the powered prosthesis platform we implement the controller on. We realize this controller on 2 subjects and present the resultant human-like joint trajectories on the prosthesis along with comparisons to 2 other tested control methods. Finally, we discuss future directions in Section \ref{sec:conclusion}.

\section{Human-Prosthesis Multi-Domain Model} \label{sec:model}
To model the multi-contact behavior of human walking, we use a multi-domain hybrid system model. To obtain the dynamics for this system, we model the human-prosthesis system as a walking robot.

\begin{figure} [t]  
\centering
\includegraphics[width=1\columnwidth]{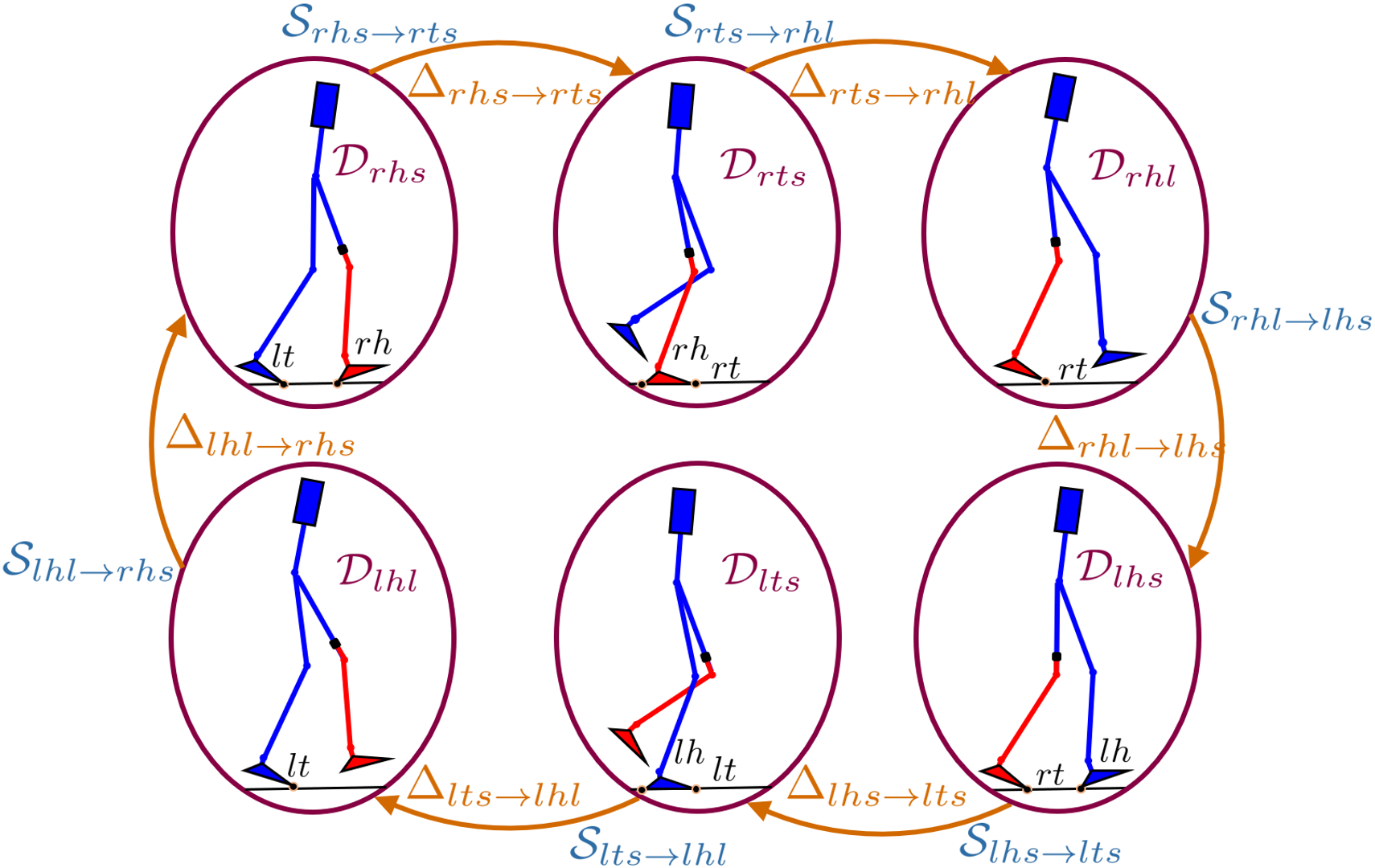}
\vspace{-0.6cm}
\caption{Six-domain directed graph of human-prosthesis hybrid system, modeling respective foot contact points for different phases of walking.}
\label{fig:hybrid-system}
\vspace{-0.6cm}
\end{figure}

\newsec{Multi-Domain Hybrid System.}
We consider 3 phases per step for human walking: heel strike (\textit{hs}) when the swinging foot's heel reaches the ground, toe strike (\textit{ts}) when that foot's toe reaches the ground, and heel lift (\textit{hl}) when the other foot's heel lifts off of the ground and becomes the swinging leg. We omit a fourth phase of the toe lifting between toe strike and heel lift because it is a very short phase. 
Since a human walking with a prosthesis is an asymmetric system, we consider separate phases for the right and left leg, prefacing the abbreviations with ``\textit{r}" and ``\textit{l}" respectively, giving a total of 6 phases. The current phase of the system is dictated by the foot contacts present. The set of all contact points is given by $\mathcal{C} = \{rh, rt, lh, lt\}$ signifying the right heel, right toe, left heel, and left toe. Fig. \ref{fig:hybrid-system} shows the contact points present for each phase, or \textit{domain}, of walking.

To account for the impact dynamics that occur when a foot strikes the ground in walking, we model the human-prosthesis system as a hybrid system. 
This \textit{multi-domain hybrid control system} is defined as a tuple \cite{ames2011human, zhao2017multi}:
\begin{equation*}
    \HybridControlSystem = (\DirectedGraph,\, \Domain,\, \ControlInput,\, \Guard,\, \Delta,\, FG),
\end{equation*}
\noindent where:
\begin{itemize}
    \item $\DirectedGraph = (V,\, E)$ is a \textit{directed cycle}, with vertices $V = \{rhs,\, rts,\, rhl,\, lhs,\, ltl,\, lhl\}$ and edges $E = \{e = \{v \rightarrow v^+\} \} |_{v \in V}$ where $v^+$ is the subsequent vertex of $v$ in the directed graph;
    \item $\Domain = \{\Domain_v\}|_{v \in V}$, set of \textit{domains of admissibility}, meaning a set of physically feasible states;
    \item $\ControlInput = \{u_v\}|_{v \in V}$, set of admissible \textit{control inputs};
    \item $\Guard = \{\Guard_e\}|_{e \in E}$, set of \textit{guards} or \textit{switching surfaces}, with $\Guard_e \subset \Domain_v$, that are the transition points between one domain $\Domain_v$ and the next $\Domain_{v^+}$ in the directed cycle;
    \item $\Delta = \{\Delta_e\}|_{e \in E}$, set of \textit{reset maps}, $\Delta_e : \Guard_e \subset \Domain_v \rightarrow \Domain_{v^+}$ that define the discrete transitions triggered at $\Guard_e$, giving the postimpact states of the system: $x^+ = \Delta_e(x)$, where $x \in \Domain_v$ and $x^+ \in \Domain_{v^+}$;
    \item $FG = \{(f_v,\, g_v)\}_{v \in V}$ with $(f_v,\, g_v)$ a \textit{control system} on $\Domain_v$, that defines the continuous dynamics $\dot{x} = f_v(x) + g_v(x)u_v$ for each $x \in \Domain_v$ and $u_v \in \ControlInput$.
\end{itemize}

\begin{figure} [t]  
\centering
\includegraphics[width=1\columnwidth]{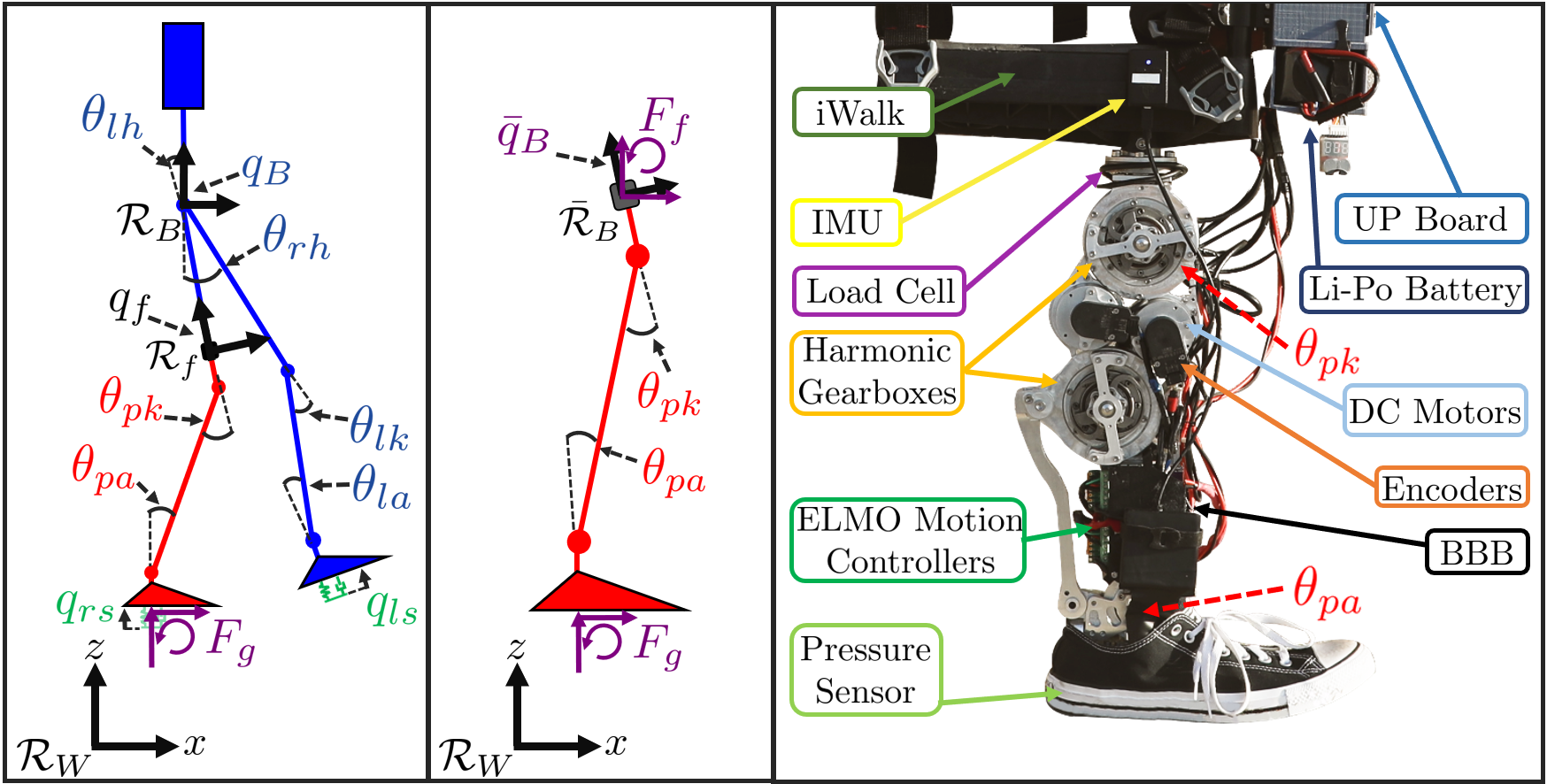}
\vspace{-0.6cm}
\caption{(left) Human-prosthesis model with the prosthesis subsystem and coordinates in red, the human remaining system and coordinates in blue, and the ground model and coordinates in green. (middle) The equivalent prosthesis subsystem modeled with its own base coordinates $\bar{q}_B$ and fixed joint forces $F_f$. (right) Powered prosthesis platform AMPRO3 labeled with subsystem coordinates and hardware component names.}
\label{fig:model-hardware}
\vspace{-0.6cm}
\end{figure}

\newsec{Human-Prosthesis Model.}
To define a control system for a human wearing a prosthesis, we model this human-prosthesis system as a series of 8 rigid links and 12 joints in 2D space, shown in Fig. \ref{fig:model-hardware}. We define these 12 DOFs by configuration coordinates $q = (q_l^T, q_f^T, q_p^T)^T \in \mathbb{R}^{12}$ which define configuration space $\mathcal{Q}$. Here $q_l$ represent the coordinates of the human, $q_f$ the 3-DOF fixed joint ($x$, $z$ Cartesian position, and pitch) that connects the human to the prosthesis at the socket, and $q_s$ represents the prosthesis subsystem coordinates.
floating base coordinates $q_B \in \mathbb{R}^3$. We have $m_r = 4$ available actuators for the human and $m_s = 2$ for the prosthesis.
The following Euler-Lagrange equation gives the constrained dynamics of this system \cite{MLS}:
\begin{align} \label{eq:robotDynamics}
    & D(q) \ddot{q} + H(q, \dot{q}) = B_v u_v + J_{h, v}^T(q) \lambda_{h, v}
    \\ \notag 
    & J_{h, v}(q) \ddot{q} + \dot{J}_{h, v}(q, \dot{q})\dot{q} = 0,
\end{align}
where $D(q)$ is the inertia matrix; $H(q, \dot{q})$ contains the centrifugal, Coriolis, and gravity terms; $B_v$ is the domain dependent actuation matrix, $u_v$ is the control input for $\Domain_v$, and $J_{h, v}(q)$ is the Jacobian of the holonomic constraints $h_v(q)$ for the contact points in $\Domain_v$. This Jacobian projects the ground reaction forces (GRFs) $\lambda_{h, v}$.

We obtain model parameters for the human portion of the system by using human inertia, limb mass, and limb length estimates based on percentage data from \cite{HumanParam, HumanInertia} and a given subject's weight, height, and sex. The prosthesis model parameters come from a CAD model of AMPRO3, the powered prosthesis used in this study \cite{zhao2017preliminary}.

\newsec{Prosthesis Subsystem.}
To implement a model-based prosthesis controller, we require a model with states and inputs that can be measured on-board. Using a floating base at the interface between the prosthesis and the human with floating base coordinates $\bar{q}_B$, we have configuration coordinates $\bar{q} = (\bar{q}_B^T, q_s^T)^T$, shown in Fig. \ref{fig:model-hardware}.
We write the dynamics for this robotic subsystem,
\begin{align} \label{eq:robotSubsystem}
    & \bar{D}(\bar{q})\ddot{\bar{q}} + \bar{H}(\bar{q}, \dot{\bar{q}}) 
    = \bar{B}_v u_{s, v} + \bar{J}^T_{h, v}(\bar{q}) \bar{\lambda}_{h, v} + \bar{J}^T_f(\bar{q}) F_f
    \\ \label{eq:holoConstrSub}
    &\bar{J}_{h, v}(\bar{q}) \ddot{\bar{q}} + \dot{\bar{J}}_{h, v}(\bar{q}, \dot{\bar{q}}) \dot{\bar{q}} = 0,
\end{align}
including the interaction forces $F_f$ it experiences with the human at the fixed joint and its projection $J_f$.
The base coordinates and their velocites $\dot{\bar{q}}$ are measured by an IMU. The interaction forces $F_f$ are measured by a load cell at the socket interface, the vertical force and pitch moment components of $\bar{\lambda}_{h, v}$ are measured by an insole pressure sensor, and the horizontal component of $\bar{\lambda}_{h, v}$ can be calculated by solving for $\ddot{\bar{q}}$ in the dynamics \eqref{eq:robotSubsystem} and substituting it into the holonomic constraint equation \eqref{eq:holoConstrSub}:
\begin{equation*} \label{eq:force}
\resizebox{0.9\hsize}{!}{$
    \bar{\lambda}_{h, v} = (\bar{J}_{h, v} \bar{D}^{-1} \bar{J}_{h, v}^T)^{-1} (\bar{J}_{h, v} \bar{D}^{-1}(\bar{H} - \bar{B}_v u_{s, v} - \bar{J}_f^T F_f) - \dot{\bar{J}}_{h, v} \dot{\bar{q}}),
    $}
\end{equation*}
 where we now drop the arguments for notational simplicity.
 
\newsec{Compliant Ground Model.}
To account for the compliance the prosthesis foot and human foot experience in ground contact with their shoes, we include a spring-damper at the base of each foot in our robot model to serve as a ``ground model". 
We use a spring stiffness of $60,000 N/m$ and a damping coefficient of $600 Ns/m$ \cite{EffectsofShoe, zhao20163d}. These prismatic joints have coordinates $q_{rs}$ and $q_{ls}$ for the right and left spring, respectively.
This allows us to generate model-based trajectories with more realistic impact dynamics.
Since on hardware the prosthesis is measuring the forces at the level of the shoe insole, instead of the forces beneath the shoe, we do not include this ground model in our prosthesis subsystem. 

\newsec{Nonlinear Control System.}
We form the full-order system robot dynamics \eqref{eq:robotDynamics} into a nonlinear control system ODE with states $x_q = (q^T, \dot{q}^T)^T $:
\begin{equation} \label{eq:roboticODE}
    \dot{x}_q = 
    \underbrace{
    \begin{bmatrix}
    \dot{q} \\
    D^{-1}(-H + J_{h, v}^T \lambda_{h, v} )
    \end{bmatrix}
    }_{f_{q, v}(x_q)}
    +
    \underbrace{
    \begin{bmatrix}
    0 \\
    D^{-1}B_v
    \end{bmatrix}
    }
    _{g_{q, v}(x_q)}
    u_v.
\end{equation}
By rearranging the states as $x = (x_r^T, x_s^T)^T$, with $x_r = (q_l^T, q_f^T, \dot{q}_l^T, \dot{q}_f^T)^T$ and $x_s = (q_s^T, \dot{q}_s^T)^T$, we obtain a control system of this form \cite{gehlhar2019control}:
\begin{align*} 
    \begin{bmatrix}
        \dot{x}_r \\ \dot{x}_s
    \end{bmatrix}
    &=
    \underbrace{\fmatrix{}}_{f_v(x)}
    + \underbrace{\gmatrix{}}_{g_v(x)}
    \begin{bmatrix}
        u_{r, v} \\ u_{s, v}
    \end{bmatrix},
    \\ \notag
        x_r \in &\mathbb{R}^{n_r}, \,\,
        x_s \in \mathbb{R}^{n_s}, \,\,
        u_{r, v} \in \mathbb{R}^{m_{r, v}}, \,\,
        u_{s, v} \in \mathbb{R}^{m_{s, v}}.
\end{align*}
We consider the bottom row of this matrix to be the \textit{separable subsystem} and the top row to be the \textit{remaining system} \cite{gehlhar2019control, StableRobustHZD}. This form shows the dynamics of $x_s$ do not depend on inputs $u_{r, v}$, allowing us to construct model-based controllers for subsystem outputs independent of the remaining system. While this subsystem depends on full-system states, we obtain an \textit{equivalent subsystem} by the same ODE transformation used for \eqref{eq:roboticODE}:
\begin{align} \label{eq:Subsystem'}
        &\dot{x}_s = \bar{f}^s(\mathcal{X}) + \bar{g}^s(\mathcal{X}) u_s,
        \\ \notag
        & \mathcal{X} = (\bar{x}_r^T,\, x_s^T,\, F_f^T)^T \in \mathbb{R}^{\bar{n}}.
\end{align}
Here $\bar{x}_r = (\bar{q}_B^T, \dot{\bar{q}}_B^T)^T$ are \textit{measurable states}
For this robotic system form, a transformation $T$ exists such that $T(x) = \mathcal{X}$ and $\bar{f}^s_v(\mathcal{X}) = f^s_v(x)$ and $\bar{g}^s_v(\mathcal{X}) = g^s_v(x)$ for all $x$ \cite{gehlhar2019control}. With an IMU and load cell to complete the model, we can now develop model-based controllers for the prosthesis subsystem based solely on the prosthesis model and locally available sensory information.

\section{Generating Subject-Specific Human-Inspired Walking Trajectories} \label{sec:trajectories}
To develop a model-based controller for the prosthesis, we start by constructing outputs for the entire human-prosthesis system such that we can generate prosthesis trajectories that are provably stable with a human model. To generate trajectories that both emulate human kinematic behavior and are provably stable for a human model walking with a powered prosthesis, we included human motion capture data in a hybrid zero dynamics \cite{westervelt2018feedback} optimization problem. This combined approach was first developed in \cite{ames2014human, zhao2017multi}, here we extend it to generate gaits with subject-specific human motion capture data and realize these walking gaits on 2 subjects through an online control Lyapunov function optimization based online controller. 

\newsec{Separable Output Construction.} We define the outputs as,
\begin{equation*} \label{eq:outputs}
    y_v(x) = y^a_v(x) - y_v^d(\tau_v(x), \alpha_v),
\end{equation*}
where $y^a_v(x)$ is the actual output of $\Domain_v$ and $y^d_v(\tau_v(x), \alpha_v)$ is the desired output given by a \Bezier polynomial defined by parameters $\alpha_v$ and modulated by a state-based phase variable: 
\begin{equation*}
    \tau_v(x) = \frac{\delta_v(x) - \delta^0_v}{\delta^f_v - \delta^0_v}.
\end{equation*}
Here $\delta_v(x)$ is a function of the states that monotonically increases from $\delta^0_v$ to $\delta^f_v$ during a domain.

To develop a controller for the subsystem \eqref{eq:Subsystem'}, we select \textit{separable outputs} $y_v(x) = (y^{r T}_v(x), y^{s T}_v(x_s) )^T$, such that outputs for the subsystem $y^s_v(x_s)$ and their Lie derivatives \cite{IsidoriNonlinSyst} do not require information about the remaining system \cite{gehlhar2019control}. By selecting outputs that are either functions of the prosthesis joints or human joints, we can define the following separable subsystem outputs from this set.
\begin{align*}
    y_{s, v}^a (x_s) |_{v \in \{rhs, rhl, lhs, ltl, lhl\} } &= [\theta_{pk},\, \theta_{pa}]^T,
    \\
    y_{s, rts}^a (x_s) &= [v_{rhip},\, \theta_{pk}]^T,
\end{align*}
with,
\begin{align*}
    v_{rhip} (x_s) &= (\bar{r}_{B} + r_{pk}) \dot{\theta}_{pk} + (\bar{r}_{B} + r_{pa}) \dot{\theta}_{pa},
\end{align*}
where $r_{\scriptscriptstyle\square}$ is the length between the joint specified in the subscript and the following distal joint, and $\bar{r}_B$ is the length between the prosthesis base frame and the prosthesis knee. 

To do state-based control for the first four domains, we define,
\begin{align*}
    \delta_{rhs} (x_s) &= \bar{\theta}_{By},
    \\
    \delta_v (x_s) |_{v \in \{ rts, rhl, lhs\} } &= \bar{r}_{B} \bar{\theta}_{By} +  r_{pk} \theta_{pk} + r_{pa} \theta_{pa}.
\end{align*}
For $\Domain_{lts}$ and $\Domain_{lhl}$ we calculate $\tau_v$ based on the current time and predicted time duration from trajectory generation.

\newsec{Human Walking Data.}
We used the average human relative joint trajectories from the motion capture data set of \cite{camargo2021comprehensive}. For each of our subjects, we selected a data set obtained with subjects with similar height and mass.
To divide the data into segments for each domain, we used 
gait cycle percentage estimates of walking phases from \cite{perry2010gait}. We used the first $12\%$ of the data points for $\Domain_{rhs}$; 
the next $19\%$ for $\Domain_{rts}$;
the next $19\%$ for $\Domain_{rtl}$;
and the final $12\%$, $19\%$, and $19\%$ for $\Domain_{lhs}$, $\Domain_{lts}$, and $\Domain_{lhl}$, respectively.
For each segment of data for a given domain, we fit \Bezier polynomials
to the data using $\mathrm{fit}$ in MATLAB. However, since there is not an impact causing discrete dynamics between the \textit{ts} and \textit{hl} domains, we have a single \Bezier stretching across both domains. Because we wanted to generate periodic gaits and the average data trajectories had a large gap between end points, we included the first data point again at the end of the data series so the \Bezier polynomials would yield periodic trajectories. 
This process gave a set of \Bezier coefficients $\alpha^H_v$ for each domain. 

\newsec{Hybrid Zero Dynamics Optimization.}
To generate desired trajectories similar to the human data \Bezier polynomials that are also provably stable for our prosthesis model, we constructed a human model for each subject using the parameters described in \ref{sec:model}. To guarantee stability for these human models, we construct a constraint for our hybrid system. In the continuous dynamics of the domains $\Domain_v$, we drive our outputs to 0 through the control input $u_v$. This reduces the system to a lower dimensional manifold called the \textit{partial hybrid zero dynamics} (PHZD) surface (or \textit{hybrid zero dynamics} surface in domains without a velocity modulating output) \cite{ames2014human}: 
\begin{equation*}
\textbf{PZ}_{\alpha_v} = \{x \in \mathcal{D}_v: \, y_{2, v}(x, \alpha_v) = 0,\, \dot{y}_{2, v}(x, \alpha_v) = 0 \}.
\end{equation*}
The control input $u_v$ only guarantees PHZD for the continuous dynamics of $\Domain_v$. To guarantee PHZD for the entire multi-domain hybrid system, we require the PHZD remain invariant through impact, mathematically expressed as,
\begin{equation*}
    \Delta_{e}(\Guard_{e} \cap \textbf{PZ}_{\alpha_{v}}) \subseteq \textbf{PZ}_{\alpha_{v^+}}.
\end{equation*}
To find parameters $\alpha_v$ that define desired trajectories that both match human data and satisfy this invariant condition, we formulate the following optimization:
\begin{align*}
    c_v^\star &= \argmin \sum_{i = 1}^6 w_i( \theta^j_i - \mathbf{B}(\alpha_{v, i}^H, \tau_v(x) )^2 + u_v^T u_v 
    \\ 
    &\quad\quad \mathrm{s.t.} \quad
    \Delta_{e}(\Guard_{e} \cap \textbf{PZ}_{\alpha_{v}}) \subseteq \textbf{PZ}_{\alpha_{v^+}}
    \\
    &\quad\quad\quad\quad \textbf{(physical constraints)}
    \\
    c_v^\star &= (\alpha_{v}^{\star T}, v_{hip, v}^{d \star}, \delta_v^{0 \star}, \delta_v^{f, \star})^T .
\end{align*}
Here we minimize the difference between the joint states $\theta^j = (\theta_{lh}, \theta_{lk}, \theta_{la}, \theta_{rh}, \theta_{pk}, \theta_{pa})^T$ and the \Bezier fit $\mathbf{B}(\alpha_{v, i}^H, \tau_v(x))$ of human joint data evaluated at the phase variable. We simultaneously minimize the torque to reduce the energy expenditure and generate smoother torque profiles. We include other physical constraints such as holonomic constraints, virtual constraints (outputs), and decision variable bounds. The solution to the optimization $c_v^\star$ gives parameters $\alpha_v^\star$ to define our desired trajectories for each domain, desired hip velocity $v_{hip, v}^d$, and phase variable parameters $\delta_v^{0 \star}$ and $\delta_v^{f \star}$. We solve this nonlinear hybrid optimization problem through FROST software \cite{hereid20163d}.

\section{Controller Realization} \label{sec:control}
To enforce these trajectories on the prosthesis subsystem, we employ a rapidly exponentially stabilizing control Lyapunov function (RES-CLF), following the construction method in \cite{ames2014rapidly}.

\newsec{Subsystem RES-CLF.}
We differentiate the outputs,
\begin{equation} \label{eq:linearized}
    \begin{bmatrix}
        \dot{y}_{1, v}^s \\
        \ddot{y}_{2, v}^s
    \end{bmatrix}
    =
    \underbrace{
    \begin{bmatrix}
        L_{\bar{f}^s_v} y^s_{1, v} \\
        L_{\bar{f}^s_v}^2 y^s_{2, v} 
    \end{bmatrix}
    }_{L_{\bar{f}^s_v} y^s_v}
    +
    \underbrace{
    \begin{bmatrix}
        L_{\bar{g}^s_v} y^s_{1, v} \\
        L_{\bar{g}^s_v}L_{\bar{f}^s_v} y^s_{2, v} 
    \end{bmatrix}
    }_{\bar{A}_v}
    u_s,
\end{equation}
with Lie derivatives \cite{IsidoriNonlinSyst} in $L_{\bar{f}^s_v} y^s_v$ and $\bar{A}_v$. Here $\bar{A}_v$ is invertible because the outputs are linearly independent, making the system feedback linearizable \cite{IsidoriNonlinSyst} with the control law,
\begin{equation} \label{eq:feedlin}
    u_s(\mathcal{X}) = \bar{A}_v^{-1} (-L_{\bar{f}^s_v} y^s_v + \nu),
\end{equation}
where $\nu$ is our auxiliary control law a user can select to stabilize the linearized system which takes this form with coordinates $\xi = (y^{s T}_1, y^{s T}_2, \dot{y}^{s T}_2)^T$,
\begin{equation*}
    \dot{\xi} = 
    \underbrace{
    \begin{bmatrix}
        0 & 0 & 0 \\
        0 & 0 & \textbf{I} \\
        0 & 0 & 0
    \end{bmatrix}
    }_F
    \xi
    \underbrace{
    \begin{bmatrix}
        \textbf{I} & 0
        \\ 
        0 & 0
        \\ 
        0 & \textbf{I}
    \end{bmatrix}
    }_G
    \nu .
\end{equation*}
For this linear system and weighting matrix $Q = Q^T > 1$, we solve the continuous time algebraic Riccati equation for solution $P = P^T > 0$ to construct a RES-CLF:
\begin{equation*}
    V(\xi) = \xi^T 
    \underbrace{
    \textbf{I}_\varepsilon P \textbf{I}_\varepsilon
    }_{P_\varepsilon}
    \xi, 
    \quad
    \mathrm{with}\,\, 
    \textbf{I}_\varepsilon := \mathrm{diag}(\textbf{I}, \frac{1}{\varepsilon} \textbf{I}, \textbf{I}).
\end{equation*}
Taking the derivative yields the convergence constraint:
\begin{equation*}
    \dot{V}(\xi, \nu) = L_F V(\xi) + L_G V(\xi) \nu \leq - \frac{1}{\varepsilon} \frac{\lambda_{min} (Q)}{\lambda_{max} (P)} V(\xi),
\end{equation*}
with Lie derivatives along the linearized output dynamics,
\begin{align*}
L_F V(\xi) &= \xi^T (F^T P_\varepsilon + P_\varepsilon F) \xi,
\\ 
L_G V(\xi) &= 2 \xi^T P_\varepsilon G.
\end{align*}

\newsec{Force Sensing ID-CLF-QP.}
To develop a hardware implementable form of a RES-CLF, we construct a variation of the inverse dynamics CLF quadratic program (ID-CLF-QP), introduced in \cite{reher2020inverse}, that was developed for the prosthesis subsystem in \cite{gehlhar2022powered}. To prescribe a desired behavior to our output dynamics $(\dot{y}_{1, v}^{s T}, \ddot{y}_{2, v}^{s T}) = \nu$, we define a desired auxiliary control input, 
\begin{equation*}
    \nu_{\rm{pd}} := K_p y^s_{2, v}(x_s) + K_d \dot{y}^s_{2, v}(x_s) + K_{y^a} \dot{y}^{s, a}_{2, v}(x_s) + K_v y^s_{1, v}(x_s).
\end{equation*}
We added the $K_{y^a} \dot{y}^{s, a}_{2, v}(x_s)$ to the typical output PD law used \cite{reher2020inverse, ames2014rapidly} to reduce oscillations observed on hardware. In our QP cost we will minimize the difference between our actual auxiliary control input $\nu$ and $\nu_{\rm{pd}}$. If we solved \eqref{eq:feedlin} for $\nu$, the expression would involve computationally expensive matrix inversions prone to numerical error \cite{reher2020inverse}. Instead we use $\nu = (\dot{y}^{s T}_{1, v}, \ddot{y}^{s T}_{2, v})^T$ and rewrite our outputs in terms of the subsystem configuration coordinates $\bar{q}$ and velocities $\dot{\bar{q}}$:
\begin{equation*} \label{eq:yddot}
    \begin{bmatrix}
        \dot{y}^{s}_{1, v} \\
        \ddot{y}^{s}_{2, v}
    \end{bmatrix}
    =
    \underbrace{
    \begin{bmatrix}
        \frac{\partial y^s_{1, v}}{\partial \bar{q}}
        \\
        \frac{\partial}{\partial \bar{q}} \bigg( \frac{\partial y^s_{2, v}}{\partial \bar{q}} \dot{\bar{q}} \bigg)
    \end{bmatrix}
    }_{\dot{J}_y(\bar{q}, \dot{\bar{q}})} \dot{\bar{q}} 
    + 
    \underbrace{ 
    \begin{bmatrix}
        \frac{\partial y^s_{1, v}}{\partial \dot{\bar{q}}}
        \\
        \frac{\partial y_s}{\partial \bar{q}}
    \end{bmatrix}
    }_{J_y(\bar{q})} \ddot{\bar{q}}.
\end{equation*}
We will include these terms in the QP cost with the holonomic constraints, enforcing these as soft constraints, using,
\begin{align*}
    &J_c(\bar{q}) = 
    \begin{bmatrix}
    J_y(\bar{q}) \\
    \bar{J}_h(\bar{q})
    \end{bmatrix},
    &\dot{J}_c(\bar{q}, \dot{\bar{q}}) =
    \begin{bmatrix}
    \dot{J}_y(\bar{q}, \dot{\bar{q}}) \\
    \dot{\bar{J}}_h(\bar{q}, \dot{\bar{q}})
    \end{bmatrix}.
\end{align*}
With these terms, we formulate our ID-CLF-QP,
\begin{equation} \label{eq:ID-CLF-QP+Ff+Fg}
\begin{aligned}
\Upsilon^\star_v 
= \mathop {\argmin }\limits_{{\Upsilon \in \mathbb{R}^{\eta_v}}} \,
& \Big|\Big| \dot{J}_{c, v}(\bar{q}, \dot{\bar{q}}) \dot{\bar{q}} + J_{c, v}(\bar{q}) \ddot{\bar{q}} - \mu \Big|\Big|^2 
+ \sigma W(\Upsilon) 
+ \rho \zeta
 \\ 
\textrm{s.t.} \,\, 
& \bar{D}(\bar{q})\ddot{\bar{q}} + \bar{H}(\bar{q}, \dot{\bar{q}}) 
= \bar{B}_v u_{s, v} + \bar{J}^T_{h, v}(\bar{q}) \tilde{F}_{g, v} 
\\
& \quad\quad\quad\quad\quad\quad\quad 
+ \bar{J}^T_f(\bar{q}) F_f
\\ 
& L_FV_v(\mathcal{X}) + L_GV_v(\mathcal{X}) \big(\dot{J}_{y, v} \dot{\bar{q}} + J_{y, v} \ddot{\bar{q}}\big) 
\\
& \quad\quad\quad\quad\quad\quad\quad 
\leq - \frac{\gamma}{\varepsilon} V_v(\mathcal{X}) + \rho
\\
& -u_{\rm{max}} \,\, \leq \,\,\, u_{s, v} \,\,\, \leq \,\, u_{\rm{max}}, 
\end{aligned}
\end{equation}
with decision variables $\tilde{\Upsilon}_v = (\ddot{\bar{q}}^T, u_{s, v}^T, \bar{\lambda}_{h, x, v}, \zeta)^T$.
Here $\mu^{\rm{pd}} = (\nu_{\rm{pd}}^T, 0^T )^T$, $W(\Upsilon)$ is a regularization term to make the system well-posed, $\sigma$ and $\rho$ are user-selected weights, and $\zeta$ is a relaxation term to allow the torque bounds $(-u_{\rm{max}}, u_{\rm{max}})$ to be met. (We omit the arguments of $J_y, \dot{J}_y$ for notational simplicity.) The GRFs $\tilde{F}_{g, v}$ contain the GRFs present for $\Domain_v$. We obtain the vertical GRF $F_{g, z}$ and pitch moment $M_{g, y}$ from a pressure sensor and the QP solves for x-component of the holonomic constraint $\bar{\lambda}_{h, x}$.
Even though the desired auxiliary control law $\nu_{\rm{pd}}$ differs from that which guarantees stability of the linearized system \eqref{eq:linearized}, we still have stability guarantees since the QP selects values that satisfy the CLF constraint.
The dimensions and components of this controller are domain-dependent since the outputs change with domain as well as the contact points, which changes the type of GRFs applied and computed.

\section{Experimental Results} \label{sec:experiment}
We realize this model and force-based multi-domain controller on our powered prosthesis platform for 2 subjects, resulting in human-like multi-contact behavior.

\newsec{Prosthesis Platform.}
We implement the ID-CLF-QP \eqref{eq:ID-CLF-QP+Ff+Fg} on the powered prosthesis platform AMPRO3 \cite{zhao2017preliminary}, Fig. \ref{fig:model-hardware}. This platform consists of two brushless DC motors (MOOG BN23) coupled with pulley systems and harmonic gear boxes. These motors are controlled with ELMO motion controllers (Gold Solo Whistle) which obtain joint position and velocity measurements from incremental encoders. The motor controllers send this feedback to the microprocessor, a Beaglebone Black Rev C (BBB), which returns a commanded torque. The controller algorithm is coded in C++ and ROS and runs at 111 Hz on the BBB. 

A SensorProd Inc. Tactilus Insole Sensor System, High-Performance V Series (SP049) is placed inside the prosthesis shoe to measure GRFs. An UP Board (2/32) scans the pressure sensors readings and sends the measurements to the BBB. A 6-axis load cell (M3564F, Sunrise Instruments) connects the iWalk human adapter to the proximal end of the prosthesis. The load cell signals go through its signal conditioning box (M8131) and then to the BBB. A Yost Labs 3-Space™ Sensor USB/RS232 IMU is mounted to the iWalk to measure the global rotation and velocity of the human's thigh and connects to the BBB. Everything is powered by a Zippy 4000mAh 10S LiPo battery pack. The whole system weighs 10.54 kg, while the prosthesis on its own weighs 5.95kg. For more details on the platform, see \cite{zhao2017preliminary}, and for the force sensors and IMU, \cite{gehlhar2022powered}.

\newsec{Experimental Set-up.}
Two non-amputee subjects wore the prosthesis device through an iWalk adapter. Subject 1 was a 1.7 m, 66 kg female and Subject 2 was a 1.8 m 75kg male. Both subjects wore a shoe lift on their left leg to even out the limb length difference when wearing the prosthesis. The subjects were allowed a chance to walk with the device before data recording started. Then they walked with each of the 3 controllers for at least 4 sets of 8 step cycles, taking a short break between controllers. The controllers included the ID-CLF-QP with force sensing \eqref{eq:ID-CLF-QP+Ff+Fg} (``sensor" controller), the ID-CLF-QP with no force sensing (``no sensor" controller), and a PD controller \footnote{A small mechanical issue was present in the PD controller test with Subject 2. Because the tracking results were comparable to the PD controller test with Subject 1 we consider the effect of the mechanical issue to be minor and included the results for completeness.}. When no force sensing was used, all of the GRFs were calculated through the holonomic constraints in the ID-CLF-QP. The weights and regularization terms in the ID-CLF-QP and the gains in $\nu_{\rm{pd}}$ were kept consistent for all tests. We set $v^d_{hip, v} = 0$ to allow the human to dictate the velocity of their stance progression instead of prescribing a set velocity.

\begin{table}[t]
\caption{Tracking RMSE of 3 Controllers for 2 Subjects \label{rmse}}
\vspace{-0.2cm}
\setlength{\tabcolsep}{6pt}
\begin{tabular}{l|ll|ll|}
\cline{2-5}
                                 & \multicolumn{2}{l|}{Knee RMSE (rad)} & \multicolumn{2}{l|}{Ankle RMSE (rad)} \\ \cline{2-5} 
                                 & Subject 1        & Subject 2        & Subject 1        & Subject 2       \\ \hline
\multicolumn{1}{|l|}{Sensor}     & 0.0228           & 0.0230           & 0.0179           & 0.0150          \\
\multicolumn{1}{|l|}{No Sensor}  & 0.0334           & 0.0315           & 0.0270           & 0.0306          \\
\multicolumn{1}{|l|}{PD Control} & 0.0242           & 0.0250           & 0.0494           & 0.0316          \\ \hline 
\end{tabular}%
\label{tab:subjects}
\vspace{-0.6cm}
\end{table}

\begin{figure} [t]  
\centering
\includegraphics[width=1\columnwidth]{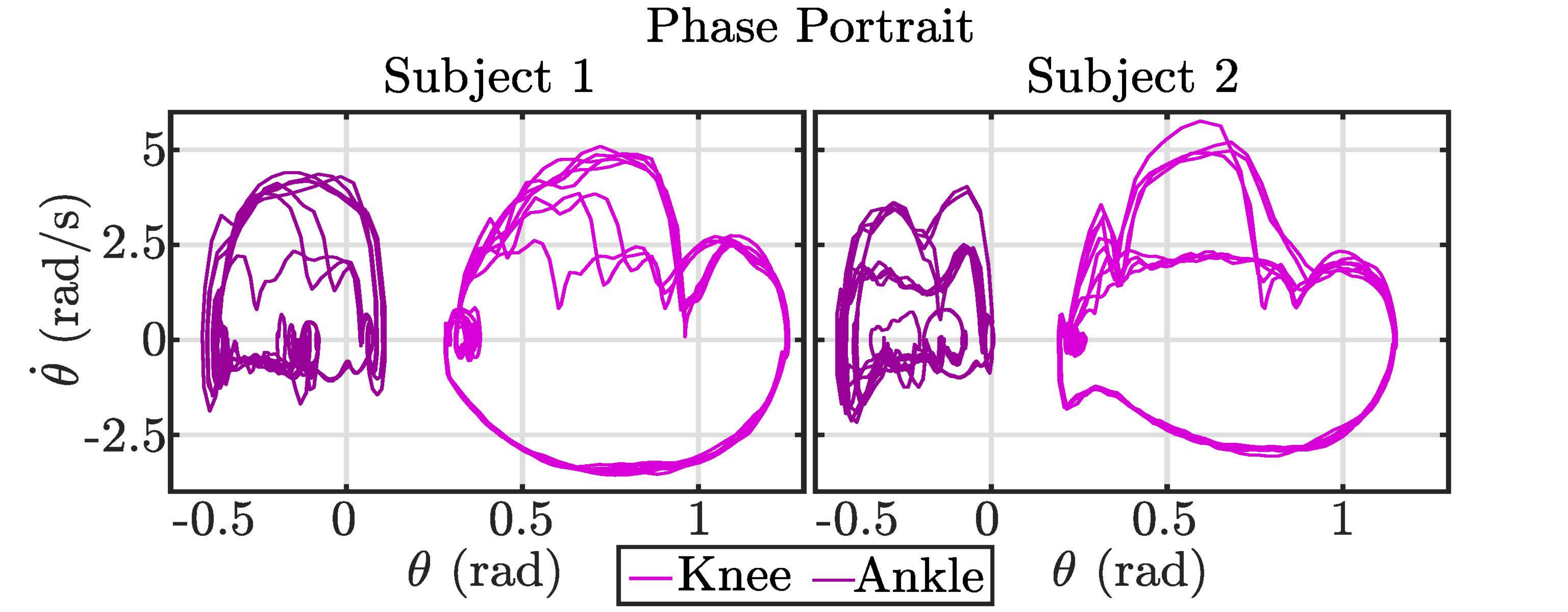}
\vspace{-0.6cm}
\caption{Phase portraits of the knee and ankle for 2 subjects for 8 continuous steps cycles using the ID-CLF-QP.}
\label{fig:phase_portraits}
\vspace{-0.2cm}
\end{figure}

\begin{figure} [t!]  
\centering
\includegraphics[width=1\columnwidth]{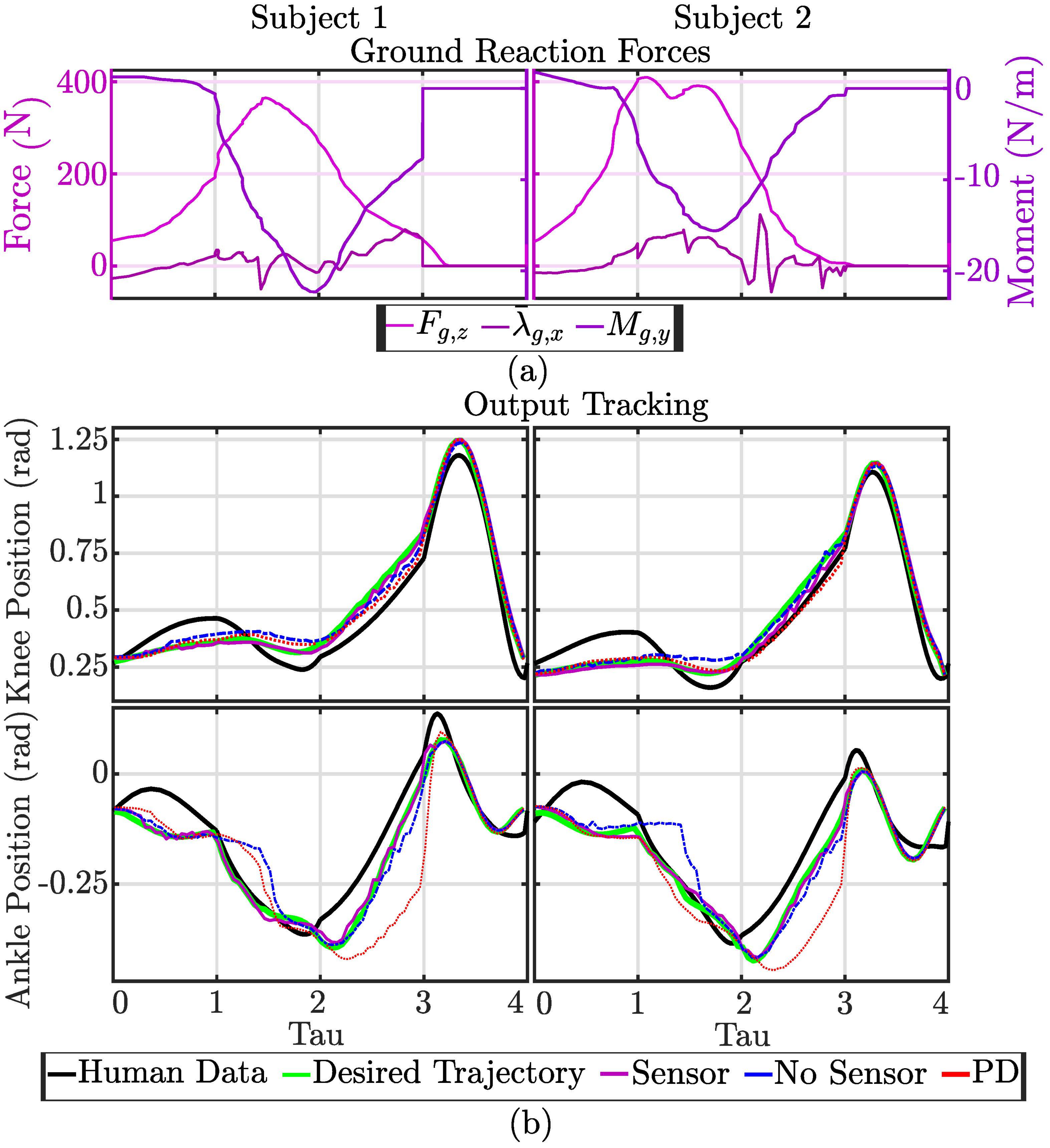}
\vspace{-0.8cm}
\caption{(a) The vertical GRF $F_{g, z}$ and moment $M_{g, y}$ measured by the pressure sensor and horizontal GRF $\bar{\lambda_{g, x}}$ during one step of walking with the ID-CLF-QP \eqref{eq:ID-CLF-QP+Ff+Fg}. (b) The mean of the actual outputs of the knee and ankle for 2 subjects and 3 controllers for 8 continuous step cycles plotted against the desired trajectory and the averaged human joint data.}
\label{fig:tracking}
\vspace{-0.4cm}
\end{figure}

\newsec{Experimental Results.}
Fig. \ref{fig:phase_portraits} shows the phase portraits of the prosthesis joints for 8 continuous step cycles with the ID-CLF-QP \eqref{eq:ID-CLF-QP+Ff+Fg}. The velocity varies between steps since it is modulated by a state-based phase variable.
Fig. \ref{fig:tracking} (a) shows the vertical GRF and pitch moment measured by the pressure sensor and horizontal GRF calculated by the QP \eqref{eq:ID-CLF-QP+Ff+Fg} during 1 step of walking with \eqref{eq:ID-CLF-QP+Ff+Fg}. 
For each controller, we computed the mean of the actual trajectories, for a set of 8 continuous step cycles, and plotted these against the desired trajectories and the subject-specific average human joint data \cite{camargo2021comprehensive} in Fig. \ref{fig:tracking} (b). Here the ID-CLF-QP with force sensing \eqref{eq:ID-CLF-QP+Ff+Fg} exhibits tight tracking to the desired trajectory, especially compared to the other controllers for the ankle. This tracking performance is quantified by the root-mean-square error (RMSE) in Table \ref{tab:subjects} showing it outperforms the other controllers for both joints and both subjects. 
More importantly, the ID-CLF-QP with force sensing matches the subject-specific human joint patterns most closely, demonstrating we can emulate this human-like behavior in a systematic way that does not involve tuning between subjects.

\section{Conclusion and Future Work} \label{sec:conclusion}
This work achieved human-like multi-contact human-prosthesis walking on 2 subjects using a model-based multi-domain controller with real-time force sensing, with no tuning between subjects. This approach provides a formally based, systematic method to generate and realize human-like motion on lower-limb powered prostheses.
In terms of tracking, this controller outperformed its counterpart without force sensors and a standard PD controller on both subjects. Being able to realize multi-contact behavior on prostheses without tuning for each subject could bring the benefits of smoother and more energy efficient gait to amputees, restoring natural and healthy locomotion.

To realize a variety of behaviors on prostheses,
gait libraries could be generated offline as in \cite{reher2021control} and in a subject-specific way using human data from a subject with similar human parameters as in this paper. This would expedite the tuning process currently required to realize human kinematic trends on powered prostheses with impedance control. 

\clearpage

\bibliographystyle{IEEEtran}
\balance
\bibliography{bibliography}

\begin{thebibliography}{10}
\providecommand{\url}[1]{#1}
\csname url@samestyle\endcsname
\providecommand{\newblock}{\relax}
\providecommand{\bibinfo}[2]{#2}
\providecommand{\BIBentrySTDinterwordspacing}{\spaceskip=0pt\relax}
\providecommand{\BIBentryALTinterwordstretchfactor}{4}
\providecommand{\BIBentryALTinterwordspacing}{\spaceskip=\fontdimen2\font plus
\BIBentryALTinterwordstretchfactor\fontdimen3\font minus
  \fontdimen4\font\relax}
\providecommand{\BIBforeignlanguage}[2]{{%
\expandafter\ifx\csname l@#1\endcsname\relax
\typeout{** WARNING: IEEEtran.bst: No hyphenation pattern has been}%
\typeout{** loaded for the language `#1'. Using the pattern for}%
\typeout{** the default language instead.}%
\else
\language=\csname l@#1\endcsname
\fi
#2}}
\providecommand{\BIBdecl}{\relax}
\BIBdecl

\bibitem{soo2010mechanics}
C.~H. Soo and J.~M. Donelan, ``{Mechanics and energetics of step-to-step
  transitions isolated from human walking},'' \emph{Journal of Experimental
  Biology}, vol. 213, no.~24, pp. 4265--4271, 12 2010.

\bibitem{fuscaldi2008effects}
L.~F. Teixeira-Salmela, S.~Nadeau, M.-H. Milot, D.~Gravel, and L.~F. Requião,
  ``Effects of cadence on energy generation and absorption at lower extremity
  joints during gait,'' \emph{Clinical Biomechanics}, vol.~23, no.~6, pp.
  769--778, 2008.

\bibitem{Inman1966HumanL}
V.~T. Inman, ``Human locomotion.'' \emph{Canadian Medical Association journal},
  vol. 94 20, pp. 1047--54, 1966.

\bibitem{HumanWalking}
V.~I. Inman, H.~J. Ralston, F.~Todd, and J.~C. Lieberman, \emph{Human Walking},
  5th~ed.\hskip 1em plus 0.5em minus 0.4em\relax Boca Raton, FL, USA: Williams
  and Wilkins, 1981.

\bibitem{kuo2001energetics}
A.~D. Kuo, ``{Energetics of Actively Powered Locomotion Using the Simplest
  Walking Model },'' \emph{Journal of Biomechanical Engineering}, vol. 124,
  no.~1, pp. 113--120, 09 2001.

\bibitem{morgenroth2011effect}
D.~Morgenroth, A.~Segal, K.~Zelik, J.~Czerniecki, G.~Klute, P.~Adamczyk,
  M.~Orendurff, M.~Hahn, S.~Collins, and A.~Kuo, ``The effect of prosthetic
  push-off on mechanical loading associated with knee osteoarthritis in lower
  extremity amputees,'' \emph{Gait \& posture}, vol.~34, pp. 502--7, 07 2011.

\bibitem{devita2007muscles}
P.~DeVita, J.~Helseth, and T.~Hortobagyi, ``Muscles do more positive than
  negative work in human locomotion,'' \emph{The Journal of experimental
  biology}, vol. 210, pp. 3361--73, 10 2007.

\bibitem{WATERS1999207}
R.~L. Waters and S.~Mulroy, ``The energy expenditure of normal and pathologic
  gait,'' \emph{Gait \& Posture}, vol.~9, no.~3, pp. 207--231, 1999.

\bibitem{Gailey_2008}
R.~Gailey, ``Review of secondary physical conditions associated with lower-limb
  amputation and long-term prosthesis use,'' \emph{The Journal of
  Rehabilitation Research and Development}, vol.~45, no.~1, pp. 15--30, dec
  2008.

\bibitem{DesignControlTransProsth}
F.~Sup, A.~Bohara, and M.~Goldfarb, ``Design and control of a powered
  transfemoral prosthesis,'' \emph{The International Journal of Robotics
  Research}, vol.~27, no.~2, pp. 263--273, 2008, pMID: 19898683.

\bibitem{GoldfarbStair}
B.~E. Lawson, H.~A. Varol, A.~Huff, E.~Erdemir, and M.~Goldfarb, ``Control of
  stair ascent and descent with a powered transfemoral prosthesis,'' \emph{IEEE
  Transactions on Neural Systems and Rehabilitation Engineering}, vol.~21,
  no.~3, pp. 466--473, 2013.

\bibitem{bhakta2019impedance}
K.~Bhakta, J.~Camargo, P.~Kunapuli, L.~Childers, and A.~Young, ``{Impedance
  Control Strategies for Enhancing Sloped and Level Walking Capabilities for
  Individuals with Transfemoral Amputation Using a Powered Multi-Joint
  Prosthesis},'' \emph{Military Medicine}, vol. 185, no. Supplement 1, pp.
  490--499, 12 2019.

\bibitem{AnnSimonConfig5}
A.~Simon, K.~Ingraham, N.~Fey, S.~Finucane, R.~Lipschutz, A.~Young, and
  L.~Hargrove, ``Configuring a powered knee and ankle prosthesis for
  transfemoral amputees within five specific ambulation modes,'' \emph{PloS
  one}, vol.~9, p. e99387, 06 2014.

\bibitem{goldfarb2014robotic}
B.~E. Lawson, J.~Mitchell, D.~Truex, A.~Shultz, E.~Ledoux, and M.~Goldfarb, ``A
  robotic leg prosthesis: Design, control, and implementation,'' \emph{IEEE
  Robotics \& Automation Magazine}, vol.~21, no.~4, pp. 70--81, 2014.

\bibitem{graham2008estimating}
K.~Ziegler-Graham, E.~J. MacKenzie, P.~L. Ephraim, T.~G. Travison, and
  R.~Brookmeyer, ``Estimating the prevalence of limb loss in the united states:
  2005 to 2050,'' \emph{Archives of Physical Medicine and Rehabilitation},
  vol.~89, no.~3, pp. 422--429, 2008.

\bibitem{sreenath2011compliant}
K.~Sreenath, H.-W. Park, I.~Poulakakis, and J.~W. Grizzle, ``A compliant hybrid
  zero dynamics controller for stable, efficient and fast bipedal walking on
  mabel,'' \emph{The International Journal of Robotics Research}, vol.~30,
  no.~9, pp. 1170--1193, 2011.

\bibitem{ames2014rapidly}
A.~D. Ames, K.~Galloway, K.~Sreenath, and J.~W. Grizzle, ``Rapidly
  exponentially stabilizing control {L}yapunov functions and hybrid zero
  dynamics,'' \emph{IEEE Transactions on Automatic Control}, vol.~59, no.~4,
  pp. 876--891, 2014.

\bibitem{Reher2020algorithmic}
J.~P. Reher, A.~Hereid, S.~Kolathaya, C.~M. Hubicki, and A.~D. Ames,
  \emph{Algorithmic Foundations of Realizing Multi-Contact Locomotion on the
  Humanoid Robot DURUS}.\hskip 1em plus 0.5em minus 0.4em\relax Cham: Springer
  International Publishing, 2020, pp. 400--415.

\bibitem{westervelt2018feedback}
E.~R. Westervelt, J.~W. Grizzle, C.~Chevallereau, J.~H. Choi, and B.~Morris,
  \emph{Feedback control of dynamic bipedal robot locomotion}.\hskip 1em plus
  0.5em minus 0.4em\relax CRC press, 2018.

\bibitem{zhao2017multi}
H.~Zhao, A.~Hereid, W.-l. Ma, and A.~D. Ames, ``Multi-contact bipedal robotic
  locomotion,'' \emph{Robotica}, vol.~35, no.~5, p. 1072–1106, 2017.

\bibitem{zhao2016multi}
H.~{Zhao}, J.~{Horn}, J.~{Reher}, V.~{Paredes}, and A.~D. {Ames},
  ``Multicontact locomotion on transfemoral prostheses via hybrid system models
  and optimization-based control,'' \emph{IEEE Transactions on Automation
  Science and Engineering}, vol.~13, no.~2, pp. 502--513, April 2016.

\bibitem{galloway2015torque}
K.~Galloway, K.~Sreenath, A.~D. Ames, and J.~W. Grizzle, ``Torque saturation in
  bipedal robotic walking through control lyapunov function-based quadratic
  programs,'' \emph{IEEE Access}, vol.~3, pp. 323--332, 2015.

\bibitem{reher2020inverse}
J.~{Reher}, C.~{Kann}, and A.~D. {Ames}, ``An inverse dynamics approach to
  control lyapunov functions,'' in \emph{2020 American Control Conference
  (ACC)}, 2020, pp. 2444--2451.

\bibitem{gehlhar2019control}
R.~Gehlhar, J.~Reher, and A.~D. Ames, ``Control of separable subsystems with
  application to prostheses,'' \emph{arXiv preprint arXiv:1909.03102}, 2019.

\bibitem{StableRobustHZD}
A.~E. {Martin} and R.~D. {Gregg}, ``Stable, robust hybrid zero dynamics control
  of powered lower-limb prostheses,'' \emph{IEEE Transactions on Automatic
  Control}, vol.~62, no.~8, pp. 3930--3942, 2017.

\bibitem{gehlhar2021separable}
R.~Gehlhar and A.~D. Ames, ``Separable control lyapunov functions with
  application to prostheses,'' \emph{IEEE Control Systems Letters}, vol.~5,
  no.~2, pp. 559--564, 2021.

\bibitem{gehlhar2022powered}
R.~Gehlhar, J.-h. Yang, and A.~D. Ames, ``Powered prosthesis locomotion on
  varying terrains: Model-dependent control with real-time force sensing,''
  \emph{IEEE Robotics and Automation Letters}, vol.~7, no.~2, pp. 5151--5158,
  2022.

\bibitem{ames2011human}
A.~D. Ames, R.~Vasudevan, and R.~Bajcsy, ``Human-data based cost of bipedal
  robotic walking,'' in \emph{Proceedings of the 14th International Conference
  on Hybrid Systems: Computation and Control}, ser. HSCC '11.\hskip 1em plus
  0.5em minus 0.4em\relax New York, NY, USA: Association for Computing
  Machinery, 2011, p. 153–162.

\bibitem{MLS}
R.~M. Murray, S.~S. Sastry, and L.~Zexiang, \emph{A Mathematical Introduction
  to Robotic Manipulation}, 1st~ed.\hskip 1em plus 0.5em minus 0.4em\relax Boca
  Raton, FL, USA: CRC Press, Inc., 1994.

\bibitem{HumanParam}
S.~Plagenhoef, F.~G. Evans, and T.~Abdelnour, ``Anatomical data for analyzing
  human motion,'' \emph{Research Quarterly for Exercise and Sport}, vol.~54,
  no.~2, pp. 169--178, 1983.

\bibitem{HumanInertia}
W.~Erdmann, ``Geometry and inertia of the human body - review of research,''
  \emph{Acta of Bioengineering and Biomechanics}, vol.~1, pp. 23--35, 1999.

\bibitem{zhao2017preliminary}
H.~Zhao, E.~Ambrose, and A.~D. Ames, ``Preliminary results on energy efficient
  {3D} prosthetic walking with a powered compliant transfemoral prosthesis,''
  in \emph{Robotics and Automation (ICRA), 2017 IEEE International Conference
  on}.\hskip 1em plus 0.5em minus 0.4em\relax IEEE, 2017, pp. 1140--1147.

\bibitem{EffectsofShoe}
G.~Andréasson and L.~Peterson, ``Effects of shoe and surface characteristics
  on lower limb injuries in sports,'' \emph{International Journal of Sport
  Biomechanics}, vol.~2, no.~3, pp. 202 -- 209, 1986.

\bibitem{zhao20163d}
H.~Zhao, A.~Hereid, E.~Ambrose, and A.~D. Ames, ``3d multi-contact gait design
  for prostheses: Hybrid system models, virtual constraints and two-step direct
  collocation,'' in \emph{2016 IEEE 55th Conference on Decision and Control
  (CDC)}, 2016, pp. 3668--3674.

\bibitem{ames2014human}
A.~D. Ames, ``Human-inspired control of bipedal walking robots,'' \emph{IEEE
  Transactions on Automatic Control}, vol.~59, no.~5, pp. 1115--1130, 2014.

\bibitem{IsidoriNonlinSyst}
A.~Isidori, \emph{Nonlinear Control Systems}.\hskip 1em plus 0.5em minus
  0.4em\relax Springer London, 1995.

\bibitem{camargo2021comprehensive}
J.~Camargo, A.~Ramanathan, W.~Flanagan, and A.~Young, ``A comprehensive,
  open-source dataset of lower limb biomechanics in multiple conditions of
  stairs, ramps, and level-ground ambulation and transitions,'' \emph{Journal
  of Biomechanics}, p. 110320, 2021.

\bibitem{perry2010gait}
J.~Perry and J.~M. Burnfield, \emph{Gait Analysis: Normal and Pathological
  Function}, 2nd~ed.\hskip 1em plus 0.5em minus 0.4em\relax New York, NY, USA:
  Slack Incorporated, 2010.

\bibitem{hereid20163d}
A.~Hereid, E.~A. Cousineau, C.~M. Hubicki, and A.~D. Ames, ``{3D} dynamic
  walking with underactuated humanoid robots: A direct collocation framework
  for optimizing hybrid zero dynamics,'' in \emph{Robotics and Automation
  (ICRA), 2016 IEEE International Conference on}.\hskip 1em plus 0.5em minus
  0.4em\relax IEEE, 2016, pp. 1447--1454.

\bibitem{reher2021control}
J.~{Reher} and A.~D. {Ames}, ``A control lyapunov function-based controller for
  compliant hybrid zero dynamic walking on cassie,'' \emph{submitted to IEEE
  Transactions on Robotics (T-RO)}, 2021 (In Preparation).

\end{thebibliography}

\end{document}